\documentclass[letterpaper, 10pt, conference]{ieeeconf}
\pdfoutput=1
\IEEEoverridecommandlockouts
\usepackage[utf8]{inputenc}
\usepackage[T1]{fontenc}
\usepackage{algorithm}
\usepackage[noend]{algpseudocode}
\usepackage{amssymb}
\usepackage{amsmath}
\usepackage{array}
\usepackage{balance}
\usepackage{graphics}
\usepackage{epsfig}
\usepackage{mathptmx}
\usepackage{times}
\usepackage{color}
\usepackage{balance}
\usepackage{tabularx,ragged2e}
\usepackage{booktabs}
\usepackage{etoolbox}
\makeatletter
\patchcmd{\@makecaption}
  {\scshape}
  {}
  {}
  {}
\makeatother

\newcommand{\rpm}{\sbox0{$1$}\sbox2{$\scriptstyle\pm$}
  \raise\dimexpr(\ht0-\ht2)/2\relax\box2 }

\newcommand{\hide}[1]{---HIDDEN---}
\renewcommand{\hide}[1]{#1} 

\DeclareMathAlphabet{\mathcal}{OMS}{cmsy}{m}{n}
\DeclareMathOperator*{\argmin}{arg\,min}

\newcolumntype{C}[1]{>{\centering\let\newline\\\arraybackslash\hspace{0pt}}m{#1}}

\title{\LARGE \bf
Grasp Learning by Sampling from Demonstration
}
\author{\hide{Philipp Zech and Justus Piater}%
\thanks{The research leading to these results has received funding from the \hide{European Community's Seventh Framework
        Programme FP7/2007--2013 (Specific Programme Cooperation, Theme 3, Information and Communication Technologies) under
        grant agreement no.~610532, SQUIRREL}.}%
\thanks{\hide{Philipp Zech and Justus Piater are with the
        Faculty of Mathematics, Computer Science and Physics,
        University of Innsbruck, 6020 Innsbruck, Austria
        \texttt{philipp.zech@uibk.ac.at} and \texttt{justus.piater@uibk.ac.at}}}%
}

\begin{document}

\maketitle
\thispagestyle{empty}
\pagestyle{empty}

\begin{abstract}
Robotic grasping traditionally relies on object features or shape information
for learning new or applying already learned grasps. We argue however that
such a strong reliance on object geometric information renders grasping
and grasp learning a difficult task in the event of cluttered environments
with high uncertainty where reasonable object models are not available.
This being so, in this paper we thus investigate the application of model-free
stochastic optimization for grasp learning. For this, our proposed learning
method requires just a handful of user-demonstrated grasps and an initial prior
by a rough sketch of an object's grasp affordance density, yet no object
geometric knowledge except for its pose. Our experiments show promising applicability
of our proposed learning method.
\end{abstract}

\section{Introduction}\label{sec:1}

Efficiently learning successful robotic grasps is one of the key challenges
to solve for successfully exploiting robots for complex manipulation tasks.
Considering existing research, grasp learning methods can be grouped into
analytic and empirical (or data-driven) methods~\cite{bohg2014,sahbani2012}.
Balasubramanian~\cite{bala2012} showed that empirical grasp learning grounded
upon Programming by Demonstration (PbD) can achieve results superior to planner
based, analytic methods.

PbD is a rather simple learning concept constructed from the idea of a robot
observing a human demonstrator to then autonomously learn manipulation skills
from its observations. In the event of grasp learning, these methods usually
rely on recording hand trajectories or postures. These are then taken as a basis
for either recognizing object and hand shapes (obviously supported by vision),
analytic computation of contact points of successful grasps, or a combination of
both to learn grasps~\cite{bohg2014}. In contrast, we propose an alternate
approach in that we sidestep both the reliance on hand trajectories or postures, and object geometric
information. Instead, we only require a few user demonstrated grasps as object-relative
6D gripper poses. From these, we then learn new grasps by sampling gripper poses relative to
a canonical object pose. This ultimately results in a grasp learning method that
requires no object geometric knowledge.

Treating a grasp as a 6D pose unlocks a key advantage compared to
shape-based and analytic methods. Learned grasps are readily applicable
to known objects by just mapping the 6D gripper pose from a canonical object pose
to the actual object pose. This requires no further knowledge than
the actual object pose. Conversely, shape-based or analytic
approaches would require either reconstruction of a shape or computation of
new contact points which may easily fail due to clutter, improper segmentation,
or missing object information. As a side note, we want to mention that in this
paper we do not address pose estimation of objects but rather assume that the
pose of an object is available.

Metropolis-Hastings (MH)~\cite{hastings1970} is a popular Markov-Chain Monte Carlo
(MCMC) sampler for approximating computationally demanding probability densities
$\pi(x)$ over a state space $\mathcal{X}$ (e.g., an object-relative gripper pose space).
At its core, it constructs a Markov chain by repeatedly drawing samples $x_{i}$
from a proposal distribution $q(x|y)$ \emph{imitating} $\pi(x)$ conditioned on $y$
with $x$, $y \in \mathcal{X}$. Proposed samples are then either accepted, i.e.,
appended to the Markov chain, with probability
\begin{equation}
  \alpha(x,y) = \min\begin{Bmatrix}
    1, \frac{\pi(y) q(x \mid y)}{\pi(x) q(y \mid x)}
  \end{Bmatrix}
  \label{eq:acc}
\end{equation}
or rejected. A Markov chain constructed in this manner satisfies both \emph{ergodicity}
and \emph{irreducibility} thus assuring that its stationary distribution,
after sufficiently many iterations, is a target probability density $\pi(x)$
sought-after. Hence, as of MH's inherent capability to optimize black-box
functions, we propose its application for active learning of grasps from
demonstration to characterize an object's unknown grasp affordance density $\pi(x)$.

In this work we introduce model-free, active learning of grasps by combining MCMC
Kameleon~\cite{sejdinovic2014} and Generalized Darting Monte Carlo (GDMC)~\cite{smini2011}
(Section~\ref{sec:4}) for characterizing an objects unknown grasp affordance density $\pi(x)$.
This requires both a rough sketch of $\pi(x)$ for the former and an initial set
of modes (a set of demonstrated grasps) of $\pi(x)$ for the latter. Given this
rough sketch MCMC Kameleon then learns an approximation of $\pi(x)$, while GDMC
nudges the proposal generating process to elliptical regions around modes of $\pi(x)$
for efficient mixing between modes. Observe that the rough sketch ultimately
biases MCMC Kameleon in terms of its exploratory behavior.

We evaluate our proposed learning method in a series of carefully designed
experiments as presented in Section~\ref{sec:5}. We conclude in Section~\ref{sec:7}
after discussing our experiments in Section~\ref{sec:6}.

\section{Related Work}\label{sec:2}

The majority of research in grasp learning from demonstration builds on recording
hand trajectories~\cite{bohg2014, sahbani2012}. Ekvall and Kragi\'c~\cite{ekvall2004,ekvall2005}
present a method that uses Hidden Markov Models for classification of a demonstrated
grasp from hand trajectories, whereas Kjellstr{\"o}m et al.~\cite{kjellstrom2008}
and Romero et al.~\cite{romero2008,romero2009}, as well as Aleotti and Caselli~\cite{aleotti2006,aleotti2007}
and Lin and Sun~\cite{lin2014} classify demonstrated grasps by a nearest neighbor
search among already demonstrated grasps. Z\"ollner et al.~\cite{zollner2001} apply
Support Vector Machines for classification of demonstrated grasps.

Instead of classifying the demonstrated grasp type and thus learning concrete grasps
for specific tasks, another idea is to focus on an object's or hand's shape during
demonstration. Li and Pollard~\cite{li2005} introduce a shape-matching algorithm that
consults a database of known hand shapes for suitably grasping an object given its
oriented point representations. Contrary, Kyota et al.~\cite{kyota2005} represent an object
by voxels to identify graspable portions. These portions later are matched against
known poses for suitably grasping an object. Herzog et al.~\cite{herzog2014} learn
gripper 6D poses of grasps which are generalized to different objects by considering
general shape templates of objects. Ekvall and Kragi\'c~\cite{ekvall2007}, and Tegin et
al.~\cite{tegin2009} extend Ekvall's and Kragi\'c's previous work~\cite{ekvall2004,ekvall2005} by considering shape
primitives which are matched to hand shapes for grasping an object. Also, Aleotti and
Caselli~\cite{aleotti2011} extended their work to detect the grasped part of the object,
thus enabling generalization of learned grasps to novel objects. Hsiao and Lozano-P\'erez~\cite{hsiao2006}
segment objects into primitive shapes to map known contact points of grasps to these
shapes. They learn contact points from human demonstration.

Yet another approach is the learning of motor skills given
trajectories of human demonstrated grasps. Do et al.~\cite{do2011} interpret a hand as a
spring-mass-damper system, where proper parameterization of this system allows forming grasps.
Kroemer et al.~\cite{kroemer2010} pursue the idea of combining active learning with
reactive control based on vision to learn efficient movement primitives for grasping
from a human demonstrator. Similarly, Pastor et al.~\cite{pastor2011} also consider
the integration of sensory feedback to improve primitive motor skills to learn predictive
models that inherently describe how things should \emph{feel} during execution of a
grasping task.

A more biologically inspired path is taken by Oztop et al.~\cite{oztop2002} by
employing a neural network resembling the mirror neuron system which is trained by a
human demonstrator for autonomously acquiring grasping skills. Hueser et al.~\cite{hueser2006}
use self-organizing maps to record trajectories which are then used to learn grasping
skills by reinforcement learning.

The work of Granville et al.~\cite{de2006} treats the grasp learning problem from a
probabilistic point of view. Given repeated demonstrations a mixture model for clustering
of grasps is established to eventually learn canonical gripper poses. Faria et al.~\cite{faria2012}
also rely on a series of demonstrations for learning grasps for establishing a probabilistic
model for a grasping task. However, they further incorporate an object centric
volumetric model to infer contact points of grasps, thus also allowing generalizing grasps
to new objects.

Detry et al.~\cite{detry2011} learn grasp affordance densities by establishing an initial
grasp affordance model for an object from early visual cues. This model then is
trained by sampling. Sweeney and Grupen~\cite{sweeney2007} establish a generative
model using an object's visual appearance as well as hand positions and orientations.
Using Gibbs sampling, new grasps then are generated from that model. Kopicki et al.~\cite{kopicki2014,kopicki2016}
propose to learn grasps by fitting a gripper's shape to an object's shape
by conjoint sampling from both a contact and a hand configuration model.

In contrast to existing related work, we only rely on an object's and a gripper's
6D pose together with a handful of demonstrated grasps for learning new grasps. Thus,
our approach is model-free as it does not rely on any further object geometric information.
Given a few demonstrated grasps, our method is capable of learning new grasps for
the demonstrated object to eventually characterize its grasp affordance density.

At this point we want to clarify again that we purposely avoid to rely on object features
or their shape. This is by virtue of the increased amount of clutter that robots are
required to deal with in their environment which induces a high degree of uncertainty
and incomplete views of the world (e.g., partial instead of fully-fledged object models).
Thus, equipping robots with learning methods that actively tackle uncertainty results
in more stable running system in a noisy world.

\section{Background}\label{sec:3}

In what follows we briefly sketch the sampling algorithms our learning method
builds upon.

\subsection{Kernel Adaptive Metropolis Hastings}\label{sec:3.1}

MCMC Kameleon as proposed by Sejdinovic et al.~\cite{sejdinovic2014} is an adaptive
MH sampler approximating highly non-linear target densities $\pi(x)$ in a
reproducing kernel Hilbert space (RKHS). During its burn-in phase, at each iteration
it obtains a subsample
$\mathbf{z} = \left\{ z_{i} \right\}_{i=1}^{n}$ of the chain history
$\left\{ x_{i} \right\}_{i=0}^{t-1}$ to update the proposal distribution
$q_{\mathbf{z}}(\cdot \mid x)$ by applying kernel PCA on $\mathbf{z}$,
resulting in a low-rank covariance operator $C_{\mathbf{z}}$.
Using $\nu^{2}C_{\mathbf{z}}$ as a covariance (where $\nu$ is a scaling parameter),
a Gaussian measure with mean $k(\cdot,y)$, i.e., $\mathcal{N}(f; k(\cdot,y),
\nu^{2}C_{\mathbf{z}})$ with $y \in \mathcal{X}$, is defined. Samples $f$ from
this measure are then used to obtain target proposals $x^{*}$.

MCMC Kameleon computes pre-images $x^{*} \in \mathcal{X}$ of $f$ by solving the
 non-convex optimization problem
\begin{equation}
  \argmin_{x\in\mathcal{X}}g(x),
\end{equation}
where
\begin{IEEEeqnarray}{rCl}
  g(x) &=& \left\Vert k\left(\cdot,x\right)-f\right\Vert _{\mathcal{H}_{k}}^{2} \\
  &=& k(x,x) - 2k(x,y) - 2 \sum_{i=1}^{n}\mathbf{\beta}_{i}\left [ k(x,z_{i})-\mu_{\mathbf{_z}}(x) \right ], \nonumber
\end{IEEEeqnarray}
$\mu_{\mathbf{z}} = \frac{1}{n}\sum_{i=1}^{n}k(\cdot,z_{i})$,
the empirical measure on $\mathbf{z}$. Then, by taking a single gradient descent
step along the cost function $g(x)$ a new target proposal $x^{*}$ is given by
\begin{equation}
  x^{*} = y - \eta \nabla_{x}g(x)\rvert_{x=y} + \xi
  \label{eqn:MCMC.opt.step}
\end{equation}
where $\mathbf{\beta}$ is a vector of coefficients, $\eta$ is the gradient step size,
and $\xi \sim \mathcal{N}(0,\gamma^{2}I)$ is an additional isotropic exploration term after
the gradient.
 The complete MCMC Kameleon algorithm then is
\begin{itemize}
  \item at iteration $t+1$
  \begin{enumerate}
    \item obtain a subsample $\mathbf{z} = \left\{ z_{i} \right\}_{i=1}^{n}$ of the chain history $\left\{ x_{i} \right\}_{i=0}^{t-1}$,
    \item sample $x^{*} \sim q_{\mathbf{z}}(\cdot \mid x_{t}) = \mathcal{N}(x_{t},\gamma^{2}I + \nu^{2}M_{\mathbf{z},x_{t}}HM_{\mathbf{z},x_{t}}^{T})$,
    \item accept $x^{*}$ with MH acceptance probability as defined in equation~\eqref{eq:acc}
  \end{enumerate}
\end{itemize}
where $M_{\mathbf{z},y}=2\eta\left[\nabla_{x}k(x,z_{1})|_{x=y},\ldots,\nabla_{x}k(x,z_{n})|_{x=y}\right]$
is the kernel gradient matrix obtained from the gradient of $g$ at $y$, $\gamma$ is a noise parameter,
and $H$ is an $n \times n$ centering matrix.

\subsection{Generalized Darting Monte-Carlo}\label{sec:3.2}

Generalized Darting Monte Carlo (GDMC)~\cite{smini2011} essentially is
an extension to classic MH samplers by equipping them with mode-hopping
capabilities. Mode-hopping behavior is beneficial for both (i) approximating
highly non-linear, multimodal targets $\pi(x)$, and (ii) to counterattack
the customary random-walk behavior of classic MH samplers by efficiently mixing
between modes.

The idea underlying GDMC is to place elliptical jump regions around known modes
of $\pi(x)$. Then, at each iteration, a local MH sampler is interrupted with probability
$P_{check}$, that is, $u_{1} > P_{check}$ where $u_{1} \sim U[0,1]$
to check whether the current state $x_{t}$ is inside a jump region. If $u_{1} < P_{check}$,
sampling continues using the local MH sampler. Otherwise, on $x_{t}$
being inside a jump region, GDMC samples another region to jump to by
\begin{equation}
  P_{i} = \frac{V_{i}}{\sum_{j} V_{j}}
\end{equation}
where $i$ and $j$ are jump region indices. $V$ denotes the n-dimensional elliptical
volume
\begin{equation}
  V = \frac{\pi^\frac{d}{2}\omega^{d}\prod_{i=0}^{d}\lambda_{i}}{\Gamma(1 + \frac{d}{2})}
\end{equation}
with $d$ the number of dimensions, $\omega$ a scaling factor, and $\lambda_{i}$ the
eigenvalues resulting from the singular value decomposition of the covariance $\Sigma$
of the Markov chain, i.e., $\Sigma = USU^{\top}$ with $S = \text{diag}(\lambda_{i})$.
We take the covariance of the Markov chain in $\mathcal{X}$ as of GDMC's design to
sample in non-feature spaces. Observe that $\pi$ in this case denotes the mathematical
constant instead of the target density $\pi(x)$. Given this newly sampled region,
GDMC then computes a new state $x_{t+1}$ using the transformation
\begin{equation}
  x_{t+1} = \mu_{x_{t+1}} - U_{x_{t+1}}S_{x_{t+1}}^{\frac{1}{2}}S_{x_{t}}^{-\frac{1}{2}}U_{x_{t}}^{\top}(x_{t} - \mu_{x_{t}})
  \label{eqn:gdmc_sample}
\end{equation}
where $\mu_{\_}$ denotes jump regions' centers (the modes), and $U$ and
$S$ again result from the singular value decomposition of the covariance $\Sigma$
of the Markov chain. GDMC accepts the jump proposal $x_{t+1}$ if $u_{2} > P_{accept}$
where $u_{2} \sim U[0,1]$ and
\begin{equation}
  P_{accept} = \min\begin{Bmatrix}
    1, \frac{n(x_{t})\pi(x_{t})}{n(x_{t+1})\pi(x_{t+1})}
    \end{Bmatrix}
\end{equation}
with $n(\cdot)$ denoting the number of jump regions that contain a state $x_{i}$.
If $x_{t}$ is outside a jump region, it is counted again, i.e., $x_{t+1} = x_{t}$.

\section{Active Learning of Grasps}\label{sec:4}

We formulate a 6D gripper pose $g$ as a dual quaternion $\hat{q} = q_{\mathrm{rot}} + \epsilon q_{\mathrm{tra}}$,
where the dual unit $\epsilon > 0$ and $\epsilon^2 = 0$. The rotational part $q_{\mathrm{rot}}$
is defined as a unit quaternion, i.e., $q_{\mathrm{rot}} = (q_{w}, q_{x}, q_{y}, q_{z})$
with $||q_{\mathrm{rot}}|| = 1$. The translational part $q_{\mathrm{tra}}$ is defined as
$q_{\mathrm{tra}} = \frac{\epsilon}{2} q_{\mathrm{rot}} q_{\mathrm{tra}}^{\prime}$
with $q_{\mathrm{tra}}^{\prime} = (0, q_{x}, q_{y}, q_{z})$, an imaginary quaternion
where the latter three components describe a translation in $\mathbb{R}^{3}$.

For each grasp, we define a quality measure by the Grasp Wrench Space
(GWS)~\cite{miller1999} denoted $\mu_{GWS}$. This measure then allows us to define
a target density $\pi(g)$ with $g \in \mathcal{X}$. Observe that $\mu_{GWS}$ defines a
valid density function as $\forall g: \mu_{GWS} \geq 0$. Further, by introducing the
normalization constant $Z$ with $Z = \sum_{i=0}^{n} \mu_{GWS}^{i}$ (where $n$ is the
number of known grasps) we have that $\frac{1}{Z}\int\pi(g)\mathrm{d}g = 1$.

As a side note we want to mention that we are well aware that the GWS may not be the
ideal grasp quality metric, however, we chose to use it as (i) we are primarily interested
in learning of feasible grasps that allow to characterize an object's grasp affordance
density, and (ii) at this point do not consider the notion of a task, which requires an
alternate grasp quality metric~\cite{bala2012}.

Our active learning method takes as an input a rough sketch of $\pi(g)$ as well as a
set of demonstrated grasps. According to Sejdinovic et al.~\cite{sejdinovic2014}
such a rough sketch to initialize MCMC Kameleon does not need to be a proper Markov
chain. Instead, it suffices if it provides good exploratory properties of the target $\pi(g)$.
We construct such a rough sketch by running a random walk MH sampler
on the object to be learned. However, we do not take the resulting Markov chain
as an initial sketch but instead the set of proposals generated during the random
walk, irrespective of whether a proposal was accepted or not. The rationale behind this is
that using a random walk MH sampler generally does not result in any learned grasps
(Section~\ref{sec:6}). Hence, the resulting Markov chain essentially would be empty
and thus not inhibit any exploratory properties of $\pi(g)$. On the other
hand, the set of proposals as generated during the random walk encapsulates enough
information regarding exploratory properties of $\pi(g)$. Thence, it suffices
as a rough sketch to initialize MCMC Kameleon.

The random walk MH sampler employed for this is constituted by a Gaussian proposal
for the position and a von-Mises-Fisher proposal for the orientation, i.e.,
\begin{equation*}
  \begin{split}
    &p^{*} \sim \mathcal{N}(x^{t}, \Sigma)\\
    &q^{*}_{\mathrm{rot}} \sim \mathcal{C}_{4}(\kappa)\exp(\kappa q_{\mathrm{rot}}^{t\;\ \top} \mathbf{x}),
  \end{split}
\end{equation*}
where $\kappa$ is a concentration parameter and $\mathbf{x}$ a p-dimensional
unit direction vector. Observe that $p^{*}$ in this case does not denote an
imaginary quaternion but rather some point $p \in \mathbb{R}^{3}$; yet it is
straight forward to convert it to one by defining $q_{\mathrm{tra}}^{\prime} = (0, p^{*})$.
We use the same probability measure as defined for MCMC Kameleon by the GWS\@.

In a real-world environment, the set of demonstrated grasps would be established
by moving the robot's gripper towards a pose, where it can grasp
the object. The gripper's position in $\mathbb{R}^{3}$ as well as its
orientation about the object in SO(3) are then recorded and treated as
a demonstrated grasp. In this work however we only study our grasp learning
methods in simulation (Section~\ref{sec:6}). Thus, we manually select points on
the object's surface to then find a grasp by optimizing the gripper's pose about
its orientation~\cite{zech2015}.

Given a rough sketch of $\pi(g)$ and a set of user demonstrated grasps, the complete
learning method then can be sketched as:
\begin{itemize}
  \item at iteration $t+1$
  \begin{enumerate}
    \item[-] attempt to perform a jump move according to the procedure as outlined in
      Section~\ref{sec:3.2},
    \item[-] otherwise, sample locally using MCMC Kameleon as outlined in
      Section~\ref{sec:3.1}.
  \end{enumerate}
\end{itemize}

\subsection{Choice of Kernel}

As briefly mentioned in Section~\ref{sec:3.1}, MCMC Kameleon requires a mean function
$k(\cdot,\cdot)$ for defining a Gaussian measure in an RKHS\@. This mean function is
defined by a valid kernel function. Generally, this is just a Gaussian kernel
$k(\cdot,\cdot)$ which in most cases delivers sufficiently good results. In our
case however, choosing a Gaussian kernel is rather self-defeating as it fails to
properly capture the distance between gripper orientations. For our implementation
we thus use a variation of the Gaussian kernel, where the Euclidean distance measure
is replaced by the transformation magnitude between two dual quaternions, i.e., the
distance and rotation that need to be applied to move from one dual quaternion
$\hat{q}$ to another dual quaternion $\hat{q}^{\prime}$~\cite{lang2015},
\begin{equation}
  k(\hat{q},\hat{q}^\prime) = \sigma^2 \exp{\left(-\frac{d_{\mathrm{mag}}(\hat{q},\hat{q}^\prime)^2}{2\ell^2}\right)}
\end{equation}
where
\begin{equation}
  d_{mag}(\hat{q},\hat{q}^\prime) = \sqrt{d_{\mathrm{arc}}(q_{\mathrm{0}}, \vec{\hat{q}}_{\mathrm{rot}})^2 + c||\vec{\hat{q}}_{\mathrm{tra}}||^2}
\end{equation}
and
\begin{equation}
  d_{\mathrm{arc}}(q,q^\prime) = \min\arccos(\langle q, \pm q^{\prime} \rangle)
\end{equation}
denoting the arc section between two quaternions. At this, $q_{\mathrm{0}} = (1,0,0,0)$
denotes the zero rotation and $\vec{\hat{q}}$ the transformation magnitude between two
dual quaternions $\hat{q}$ and $\hat{q}^{\prime}$ which is defined as $\overline{\hat{q}} \cdot \hat{q}^{\prime}$,
where $\overline{\hat{q}}$ denotes the dual quaternion conjugate. Further, $c$ is a
constant larger or equal to $0$ weighting the influence of the translation on the
transformation magnitude.

During our experiments we realized however that taking the arc length between two
rotations that are infinitesimally close together often results in numerical instabilities
in case of computing $M_{\mathbf{z},y}$. We thus decided to linearize $d_{\mathrm{arc}}$
by replacing it with the Euclidean distance between the two rotations $q_{0}$ and $\vec{\hat{q}}_{\mathrm{rot}}$
resulting in a stably running algorithm. The application of this modified Gaussian
kernel for 6D poses then results in a geometrically valid distance measure.

\section{Experimental Method}\label{sec:5}

An interesting question for our learning method is to what extent the rough sketch
of an object's grasp affordance density $\pi(g)$ biases MCMC Kameleon in both its
exploratory behavior and its learning rate. Thus, in the following we discuss and
report results that can be categorized by three levels of initial bias, viz.~\emph{impartial},
\emph{weak}, and \emph{strong}. For an impartial bias, the rough sketch contains
no valid grasps, for a weak bias only the demonstrated grasps, and for a strong bias
at least 500 valid grasp samples that were established using RobWork~\cite{ellekilde2010},
a robotics and grasp simulator. The total number of valid and invalid samples for each
rough sketch was 1000.

\begin{figure}
  \centering
  \resizebox{\columnwidth}{!}{
    \includegraphics{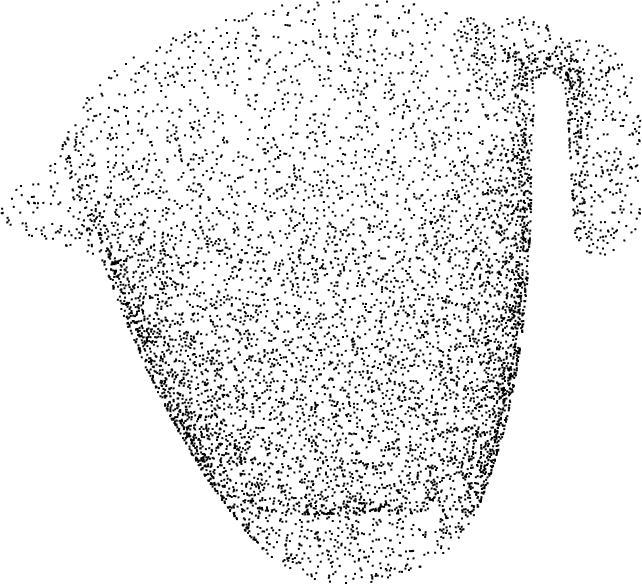}\hspace{1cm}
    \includegraphics{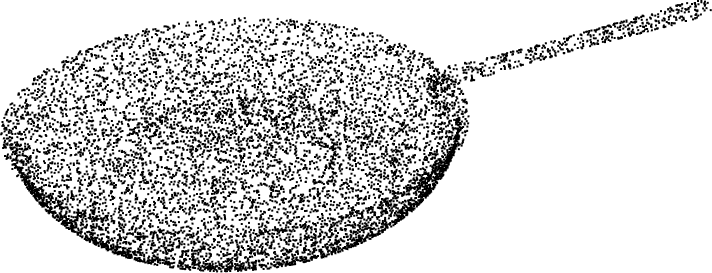}
    \includegraphics{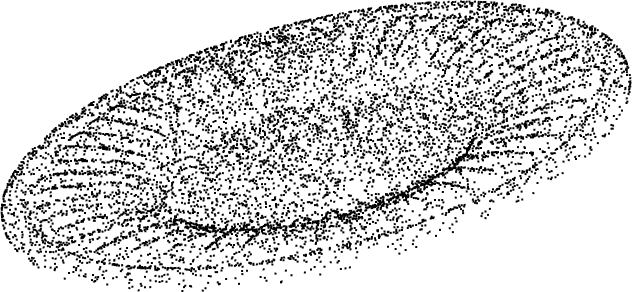}}
  \caption{Object set used for grasp learning.}
  \label{fig:objects}
\end{figure}

We evaluated our learning method on the objects depicted in Figure~\ref{fig:objects}.
In total, we did 288 simulated runs of our learning method using RobWork. Table~\ref{tab:params} s
hows our parameterization of MCMC Kameleon and GDMC for our experiments. The values
were learned during a series of preliminary experiments by bringing the acceptance rate close to $\alpha = 0.234$~\cite{andrieu2008}.
\begin{table}
  \renewcommand{\arraystretch}{1.8}
  \centering
  \caption{Parameterization of MCMC Kameleon and GDMC for our experiments.}
  \tabcolsep=0pt\def\arraystretch{1.3}
  \begin{tabularx}{240pt}{c*7{>{\Centering}X}}\toprule
    Iterations+Burn-In & $\gamma$ & $n$ & $\nu$~\cite{sejdinovic2014} & $P_{check}$ & $\omega$ & $c$ \tabularnewline \midrule
    1000+100             & 1e-05    & 100 & $\frac{2.38}{\sqrt{6}}$     & 0.5         & 0.7      & $0.0-0.2$ \tabularnewline\bottomrule
  \end{tabularx}
  \label{tab:params}
\end{table}
For all experiments we used 5 demonstrated grasps. Modulation values of $c$ for different
runs were increments of $0.005$ in the interval $[0.0-0.1]$ and $0.01$ in the interval
$[0.1-0.2]$, respectively. As mentioned in Section~\ref{sec:4} training of $C_{\mathbf{z}}$
was only done during the algorithm's burn-in to assure convergence.

We further simulated one run for each object using our random walk MH sampler as
defined in Section~\ref{sec:4} to both establish a rough sketch for MCMC Kameleon
as well as a base line to compare our proposed learning method to.

At this point we want to clarify that the need for an object model for our
experiments only arises, as we evaluate our learning method in simulation.
In the event of learning on a real robot, no object model or object geometric
information except for its pose is necessary which we assume to be available.

\section{Results and Discussion}\label{sec:6}

\begin{figure*}
  \centering
  \resizebox{\textwidth}{!}{
    \includegraphics{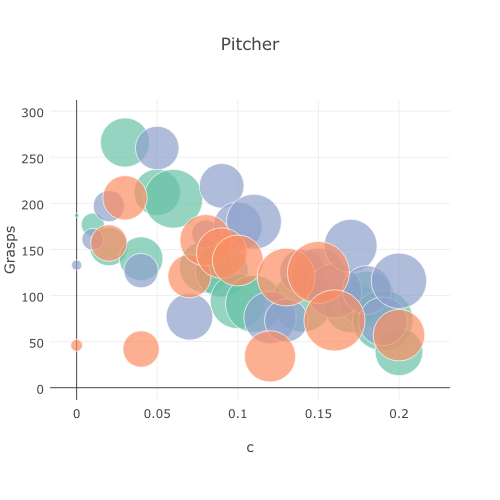}\hspace{1cm}
    \includegraphics{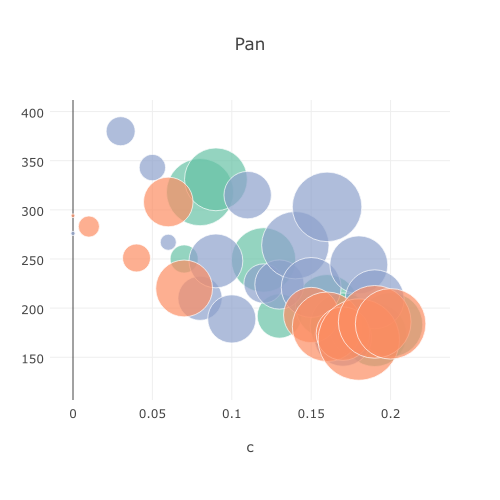}
    \includegraphics{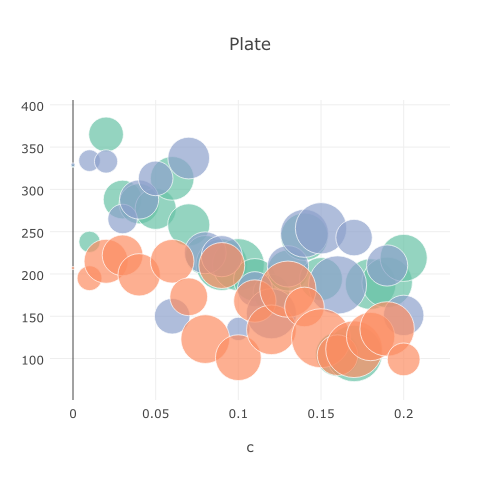}}
  \caption{Dispersion plots for all runs w.r.t.~$c$ relative to the number of successfully sampled
    grasps. The cirlces' sizes denote the surface area of the convex hull of the set of successfully
    sampled grasps, thence describing the exploratory behavior of the sampler for a given value of $c$.
    Orange denotes runs with impartial, blue with weak, and green with strong bias. Due to space restrictions
    values for $c$ were modulated with increments of $0.01$. Best viewed in color.}
  \label{fig:dispersion}
\end{figure*}

Table~\ref{tab:res_averaged} shows a general overview of the results of our experiments.
For each of the objects, the random walk MH sampler yielded, as expected, very poor
results. This is evident from the simple design of this sampler. Both the Gaussian and the
von-Mises Fisher proposal fail to properly address the high non-linearity and
multimodality of our target $\pi(g)$. Further, as of proposing position and orientation
independent of each other, the sampler also does not amount for the necessary correlation
between them. With the premise that a gripper's position constrains the reaching movement,
synergy among position and orientation is paramount~\cite{Roby2000}.
\begin{table}[h]
  \centering
  \caption{Successfully sampled grasps of our proposed learning method compared to a
    random walk MH sampler (Section~\ref{sec:4}). The values are averaged over varying $c$
    (Table~\ref{tab:params}) with the standard deviation denoted in brackets.}
    \tabcolsep=0pt\def\arraystretch{1.3}
    \begin{tabularx}{240pt}{l c*4{>{\Centering}X}}\toprule
              & Random Walk & \multicolumn{3}{c}{MCMC combined with GDMC} \tabularnewline
              &   & impartial & weak & strong \tabularnewline \midrule
      Pitcher & 0 & $143 (\rpm 48)$ & $144(\rpm 59)$ & $160(\rpm 55)$ \tabularnewline
      Pan     & 2 & $216(\rpm 46)$ & $254(\rpm 60)$ & $261(\rpm 59)$ \tabularnewline
      Plate   & 1 & $180(\rpm 46)$ & $240(\rpm 55)$ & $248(\rpm 62)$ \tabularnewline\bottomrule
    \end{tabularx}
  \label{tab:res_averaged}
\end{table}
In contrast, the results of our proposed learning method from Table~\ref{tab:res_averaged}
clearly indicate that our approach is feasible. No matter how we biased MCMC Kameleon,
our learning method, with a few exceptions, consistently found a substantial number
of grasps in the hundreds. This capitalizes on MCMC Kameleon's adaptability to its
target which allows it to learn a covariance operator $C_{\mathbf{z}}$ that describes
the relation between positions and orientations of grasps for an object. Overfitting
of $C_{\mathbf{z}}$ in terms of becoming too \emph{local} around a specific mode is
thwarted by random mode hops. Our conclusions are further strengthened by Figure~\ref{fig:dispersion}.

What is also visible from Figure~\ref{fig:dispersion} is the role of $c$. From a geometrical
point of view, the value of $c$ used in our kernel essentially constrains the gradient step
size in the translational dimensions. Thus, if set to $0$, the gradient ignores the
translation completely and only optimizes over the gripper's orientations. However, by
gradually increasing $c$ we allow the sampler to not only explore different orientations
of the gripper but also, at the same time, to move along the object to find new grasping
points in previously unexplored regions of an object. Clearly, $c$ has to be kept
at a reasonable value. Setting it too high results in that the sampler overshoots
the object by moving away from it, as of too distant moves in the translational
dimensions. From Figure~\ref{fig:dispersion} it is evident that for all three objects
ideal values for $c$ lie in the range $0.25$ to $0.1$, where in case of the pan
this may be extended to $0.15$ as of its elongated shape. Values of $c$ higher
than these upper bounds, though yielding better exploration, generally result
in less successfully sampled grasps.

Figure~\ref{fig:dispersion} also answers our initial question from Section~\ref{sec:5}. If using
an impartial bias our learning method performed poorest. Yet, the difference between using
either a weak or a strong bias is not substantial anymore. Obviously, a strong bias results
in an overfitted covariance operator $C_{\mathbf{z}}$. This again underpins our assumption
that grasp learning using only very little object knowledge by its pose and a handful of
demonstrated grasps is feasible.

\begin{figure}
  \centering
    \resizebox{\columnwidth}{!}{%
    \includegraphics{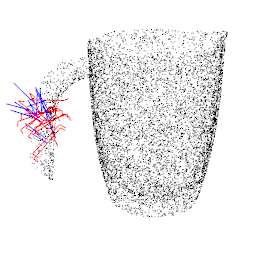}%
    \includegraphics{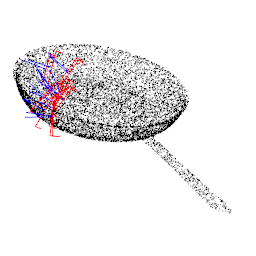}%
    \includegraphics{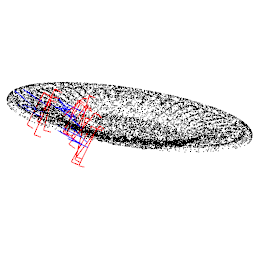}}
    \resizebox{\columnwidth}{!}{%
    \includegraphics{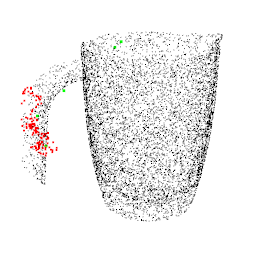}%
    \includegraphics{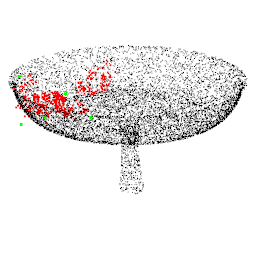}%
    \includegraphics{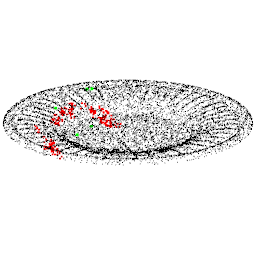}}
  \caption{Successfully sampled grasps with corresponding grasp affordance densities. Green dots
  denote the demonstrated grasps, red dots sampled grasps. For each plot, the value of $c$ is $0.08$
  with weak bias. Best viewed in color.}
  \label{fig:results}
\end{figure}

Finally, Figure~\ref{fig:results} shows for each object from Figure~\ref{fig:objects} a
few successfully sampled grasps with their corresponding (partial) grasp affordance
densities.

\section{Conclusions}\label{sec:7}

We have presented a novel method for active learning of grasps for characterizing an
object's grasp affordance density. We have shown that grasp learning is feasible
without any object knowledge except for its pose. Our learning method essentially
requires nothing more than a few demonstrated grasps.

Our approach is grounded on MCMC sampling, more specifically
a combination of MCMC Kameleon and GDMC\@. These algorithms each have advantageous
characteristics. MCMC Kameleon allows sampling from highly non-linear distributions,
whereas GDMC tackles the issue of properly exploring a multimodal distribution. We
found that a combination of both ideally fits the problem of model-free, active
learning of grasps. Our discussion of experimental results clearly corroborate our
conclusions.

We believe that the results of our work are important in that they show that
grasp learning essentially can be done \emph{blindly}. That is, by avoiding
to rely on any object geometric knowledge an object's grasp affordance denisty
can still be successfully characterized. This eventually results in more robust and
versatile robots.
\balance

\bibliographystyle{IEEEtran}
\bibliography{references}

\end{document}